\documentclass{article}

\PassOptionsToPackage{numbers}{natbib}
\usepackage[preprint]{neurips_2026}

\usepackage[preprint]{neurips_2026}

\usepackage[utf8]{inputenc} 
\usepackage[T1]{fontenc}    
\usepackage{hyperref}       
\usepackage{url}            
\usepackage{booktabs}       
\usepackage{amsfonts}       
\usepackage{nicefrac}       
\usepackage{microtype}      
\usepackage{xcolor}         
\usepackage{subcaption}

\usepackage{booktabs}
\usepackage{multirow}
\usepackage{adjustbox}

\usepackage{graphicx}
\usepackage{amsmath}
\usepackage{multirow}
\usepackage{booktabs} 
\usepackage{float}

\title{Learning Temporal Dynamics with Derivative Representations for Time Series Forecasting}

\author{%
  Shujie Duan$^{1}$, Jia Hong Puah$^{1}$, Haocheng Wu$^{1}$, Yifei Mao$^{1}$,\\
  Hongting Zhou$^{1}$, Chenxuan Li$^{1}$, Zitao Fang$^{1}$, Yang Lu$^{2}$,\\
  Sim Kuan Goh$^{1}$\thanks{Corresponding author: Sim Kuan Goh (\texttt{simkuan.goh@xmu.edu.my}).}\\[4pt]
  {\small $^{1}$School of Artificial Intelligence and Robotics,}\\
  {\small Xiamen University Malaysia, 43900 Sepang, Selangor, Malaysia}\\
  {\small $^{2}$Department of Computer Science and Technology, School of Informatics,}\\
  {\small Xiamen University, Xiamen 361005, Fujian, China}\\
  {\footnotesize \texttt{AIT2209069@xmu.edu.my}, \texttt{MEG2311003@xmu.edu.my}}\\
  {\footnotesize \texttt{AIT2409044@xmu.edu.my}, \texttt{DSC2409024@xmu.edu.my}}\\
  {\footnotesize \texttt{AIT2309094@xmu.edu.my}, \texttt{AIT2309070@xmu.edu.my}}\\
  {\footnotesize \texttt{AIT2209071@xmu.edu.my}, \texttt{luyang@xmu.edu.cn}}\\
  {\footnotesize \texttt{simkuan.goh@xmu.edu.my}}
}

\begin{document}

\maketitle

\begin{abstract}
Time series forecasting aims to predict the future evolution of complex systems by learning their underlying dynamics from past observations. Despite recent advances in deep learning, Transformers in particular still struggle to capture the temporal evolution of system dynamics. Existing approaches have been largely driven by architectural innovation, with comparatively little attention paid to training strategies that explicitly exploit the temporal structure governing system dynamics. In this work, we introduce DeltaTSF, a dynamics-aware forecasting framework based on discrete derivative representations of time series. Specifically, DeltaTSF jointly learns the series and their discrete temporal derivatives across both input and forecast horizons within a unified, model-agnostic framework. By coupling these representations over time, the model is encouraged to maintain temporal consistency, a property we show to induce a form of discrete Taylor consistency during training. This formulation enables the model to better reflect the underlying system dynamics, resulting in more stable and physically coherent forecasts. Notably, DeltaTSF does not require explicit derivative computation and introduces no additional computational overhead at inference time. Extensive experiments across diverse benchmarks and backbone architectures demonstrate that DeltaTSF consistently improves forecasting performance and yields more dynamically coherent predictions than strong SOTA methods. These results highlight the effectiveness of enforcing dynamic coherence through derivative representations as a model-agnostic paradigm that operates alongside and complements architectural design. Source code will be made publicly available upon acceptance.
\end{abstract}

\section{Introduction}

Time series forecasting underlies a wide range of real-world applications, including climate modeling~\cite{doi:10.1126/science.adi2336,wu2023interpretable}, traffic~\cite{cirstea2022towards}, energy planning~\cite{electricityloaddiagrams20112014_321}, and finance\cite{huang2024generative}, with the goal of inferring latent temporal dynamics from historical observations to accurately predict future trajectories. While deep learning models have been demonstrating outstanding performance in computer vision and natural language processing tasks, they have just started to show initial promising results when applied to time series forecasting~\cite{sun2024learning, yu2024ginar, qiu2025duet, sun2025ppgf}, and substantial challenges remain. These challenges arise from intrinsic characteristics of time series data, including non-stationarity, uncertainty, low signal-to-noise ratios, unobserved external factors, and the presence of unforeseen future events, leaving significant room for further investigation.

To address these challenges, a variety of research have been explored across multiple aspects summarized in Table~\ref{tab:taxanomy}, including time-series representations, model architectures (predominantly Transformer-based methods~\cite{zeng2023transformers,nie2023timeseriesworth64}) and its variants and backbone substitutions with other architectures, and training paradigms~\cite{kudrat2025patch}. Moreover, several studies explore approaches beyond task-specific models, including time-series foundation models pretrained on large-scale and heterogeneous temporal data~\cite{ansari2024chronos}, as well as multi-modal large language models with time-series understanding and forecasting capabilities~\cite{zhong2025time}. Despite these significant efforts, learning effective and transferable representations for time series data remains challenging. Empirical studies have reported contradictory findings; for example, ongoing debates question whether transformer-based architectures, despite their dominance in large language models, are inherently suitable for time series forecasting or require substantial modification to meet practical performance expectations~\cite{ke2024curse,zeng2023transformers}. Notably, a few studies have shown that even simple multi-layer perceptron models can outperform transformer-based models tailored for time series tasks~\cite{liang2025attention,ekambaram2023tsmixer,kim2025comprehensive}. This observation suggests that while architectural design plays an important role, other overlooked factors are equally crucial for effective time-series representation learning. Consequently, several recent works have shifted attention to model-agnostic training paradigm~\cite{kudrat2025patch} that complements architectural design.

Time series data consist of measurements of observable states from complex systems, which evolution is governed by underlying temporal dynamics. However, many existing forecasting methods primarily emphasize point-wise prediction accuracy at individual time steps~\cite{nie2023timeseriesworth64,liu2023itransformer,huangtimebase}, often overlooking the intrinsic dynamics. A line of research explicitly models dynamical state transitions, such as structured state-space models and related dynamical approaches, that do not leverage transformer architecture~\cite{chang2025deep,hu2024time}. While these methods provide clearer theoretical grounding, their practical performance on real-world forecasting datasets remains under active investigation to challenge current state-of-the-art. Nevertheless, their findings highlight the importance of capturing the dynamics of real-world systems for interpretability. Hence, we address this gap by developing a dynamics-aware training framework that implicitly encourages the learning of temporally aligned and dynamically consistent representations, and can be readily integrated into existing methods.

In this work, we study derivative representations of time series that explicitly characterise how temporal variations evolve across both input and output horizons, and propose DeltaTSF, a dynamics-aware forecasting framework that systematically exploits these representations for forecasting. Specifically, DeltaTSF jointly models series and their temporal derivatives within a shared architecture, coupling their predictions across time to enforce temporally structural alignment. We show that this coupling implicitly induces a discrete Taylor consistency across the forecast horizon during back-propagation. At inference time, DeltaTSF requires no derivative computation and produces forecasts that better respect the underlying temporal dynamics of the data. Extensive experiments across multiple forecasting benchmark datasets demonstrated that DeltaTSF consistently improves forecasting accuracy while yielding more dynamically coherent predictions than strong baselines.

\begin{table}[t]
\centering
\footnotesize
\caption{A taxonomy of recent time-series forecasting methods organized by representation space, time-series characteristics, and Transformer intervention dimensions.}
\setlength{\tabcolsep}{4pt}
\renewcommand{\arraystretch}{1.1}

\begin{adjustbox}{width=0.9\textwidth}
\begin{tabular}{lccccccc}
\toprule
\multirow{3}{*}{\textbf{Method}} 
& \multicolumn{4}{c}{\textbf{Time Series Representation}} 
& \multicolumn{3}{c}{\textbf{Transformer Intervention Dimensions}} \\
\cmidrule(lr){2-5} \cmidrule(lr){6-8}
& Raw & Decomposition & Spectral & Derivative 
& Backbone & Architectural & Training\\
& {\scriptsize (Sequential)}
& {\scriptsize (Trend/Seasonal)}
& {\scriptsize (Periodic)}
& {\scriptsize (Rate of Change)}
& Substitution & Design & Paradigm \\
\cmidrule(lr){1-1} \cmidrule(lr){2-5} \cmidrule(lr){6-8}
DLinear \cite{zeng2023transformers}      &   & $\checkmark$ &   &   & Linear   &   &   \\
PatchTST \cite{nie2023timeseriesworth64}     & $\checkmark$ &   &   &   &   & $\checkmark$ &  \\
TimesNet \cite{wu2022timesnet}     &   &   & $\checkmark$ &   & CNN      & $\checkmark$ &   \\
TimeMixer \cite{wang2024timemixer}    &   & $\checkmark$ &   &   & MLP      &   &   \\
iTransformer \cite{liu2023itransformer} & $\checkmark$ &   &   &   &          & $\checkmark$ &   \\
PatchMLP \cite{tang2025unlocking}     &   & $\checkmark$ &   &   & MLP      &   &   \\
TimeBase \cite{huangtimebase}     & $\checkmark$ &   & $\checkmark$ &   & Linear   & $\checkmark$ &   \\
PSLoss \cite{kudrat2025patch}       &   & $\checkmark$ &   &   & Agnostic &   & $\checkmark$ \\
PhaseFormer  \cite{khan2025phaseformer} &   &   & $\checkmark$ &   &          & $\checkmark$ &   \\
DeltaTST (Ours)      & $\checkmark$ &   &   & $\checkmark$ & Agnostic &   & $\checkmark$ \\
\bottomrule
\end{tabular}
\label{tab:taxanomy}
\end{adjustbox}
\end{table}

Our contributions are summarized as follows:
\begin{itemize}
    \item We propose a simple yet effective training paradigm, DeltaTSF, which learns temporally and dynamically consistent representations by coupling time series with their derivatives across both input and output horizons, which is generally applicable to any neural architecture.

    \item We show that the training process of DeltaTSF implicitly enforces a discrete Taylor consistency across the forecast horizon during backpropagation, enabling dynamically coherent forecasts, at negligible training overhead and zero additional inference cost.

    \item We conduct extensive experiments on multiple benchmark datasets and demonstrate that the proposed framework consistently improves performance when integrated into various state-of-the-art baselines. Further analysis of the training procedure and its variants highlights a promising direction for future development of dynamics-aware forecasters.

\end{itemize}

\section{Related Work}

Modeling long-range temporal dependencies for forecasting remains challenging and has attracted increasing interest from the deep learning community seeking to extend its success to time series data. Convolutional Neural Networks (CNNs) \cite{zhao2019deep,deng2019knowledge}, which rely on translational invariance and local kernels, struggle to capture long-range dependencies across historical and forecast horizons. Recurrent Neural Networks (RNNs) \cite{salinas2020deepar} and variants such as LSTM \cite{beck2024xlstm} model temporal dependencies through recurrent states, but often suffer from vanishing gradients. Hybrid CNN-RNN architectures \cite{livieris2020cnn, widiputra2021multivariate} combine the strengths of both models to mitigate these limitations with some successes. More recently, structured state-space models (SSMs) \cite{guefficiently} have provided a unifying theoretical framework for CNNs and RNNs, enabling stable modeling of long-range dependencies. In parallel, alternative frameworks such as Neural Differential Equations (NDEs) \cite{chen2018neural} incorporate ordinary or partial differential equations (ODEs/PDEs) into neural models to enforce temporal continuity and capture differential dynamics across variables and time. Despite the theoretical grounding, their practical performance on real-world time series remains underexplored, with unstable training and inference \cite{rackauckas2020universal, jia2019neural}, scaling behavior not well understood \cite{vankadara2024feature, onNDE} and suboptimal hardware utilization limiting training efficiency \cite{gu2024mamba} to challenge current state-of-the-art. Nevertheless, these findings highlight the importance of continuous dynamics and temporal consistency in real-world systems.

Following the success of large language models, transformer architectures \cite{vaswani2017attention} have been extensively studied and adapted for time series forecasting. While transformers excel in natural language processing, where tokens encode rich semantic information, time series inputs consist of numerical values that often provide only partial observations of complex systems. To address this limitation, various approaches introduce sparse attention mechanisms \cite{zhou2021informer} or condense information through patching and embedding strategies \cite{nie2023timeseriesworth64}. Specialized transformer architectures have also been proposed to model cross-variate dependencies \cite{liu2023itransformer,zhang2023crossformer,zhang2024sageformer}. However, several studies report that ignoring cross-variate dependencies entirely can yield comparable performance \cite{nie2023timeseriesworth64, huangtimebase}, highlighting ongoing uncertainty regarding architectural design choices. In parallel, there is growing interest in time series foundation models (FMs) \cite{rasul2023lag, ansari2024chronos}, which are pretrained on large-scale and heterogeneous temporal data and subsequently fine-tuned for downstream tasks. Although these models demonstrate promising initial results, limited performance scaling has been observed with increasing model size and dataset scale. Furthermore, there is ongoing debate regarding whether transformer-based models offer consistent advantages over simpler multi-layer perceptron (MLP) approaches \cite{yadav2025attention,ke2024curse,zeng2023transformers}.

Beyond architectural advances, another line of research focuses on training paradigms. The recent rise of diffusion models  (DM)~\cite{ho2020denoising} has also introduced probabilistic perspectives to time series forecasting. These probabilistic models learn the distribution of future sequences conditioned on partial observations, enabling applications such as time series forecasting \cite{kollovieh2023predict}, imputation \cite{tashiro2021csdi}, and data completion \cite{alcaraz2022diffusion}. However, DMs rely on iterative denoising processes, which result in high computational cost and slow inference. A few works introduce auxilatary loss functions for decompositional forecasting~\cite{qiudbloss}, patch-wise structural similarity~\cite{kudrat2025patch}, and self-supervised contrastive learning~\cite{park2024self}. Nonetheless, most of the aforementioned methods primarily emphasize point-wise forecasting accuracy at individual time steps, or global fidelity in the spectral domain, while largely overlooking the preservation of underlying temporal dynamics and cross-horizon consistency. Consequently, it remains unclear how temporal dynamics can be systematically and effectively incorporated into training frameworks for time-series representation learning.

\section{Method}
\label{sec:method}

\begin{figure*}[t]
  \vskip 0.2in
   \centering
    \centerline{\includegraphics[width=1.2\textwidth]{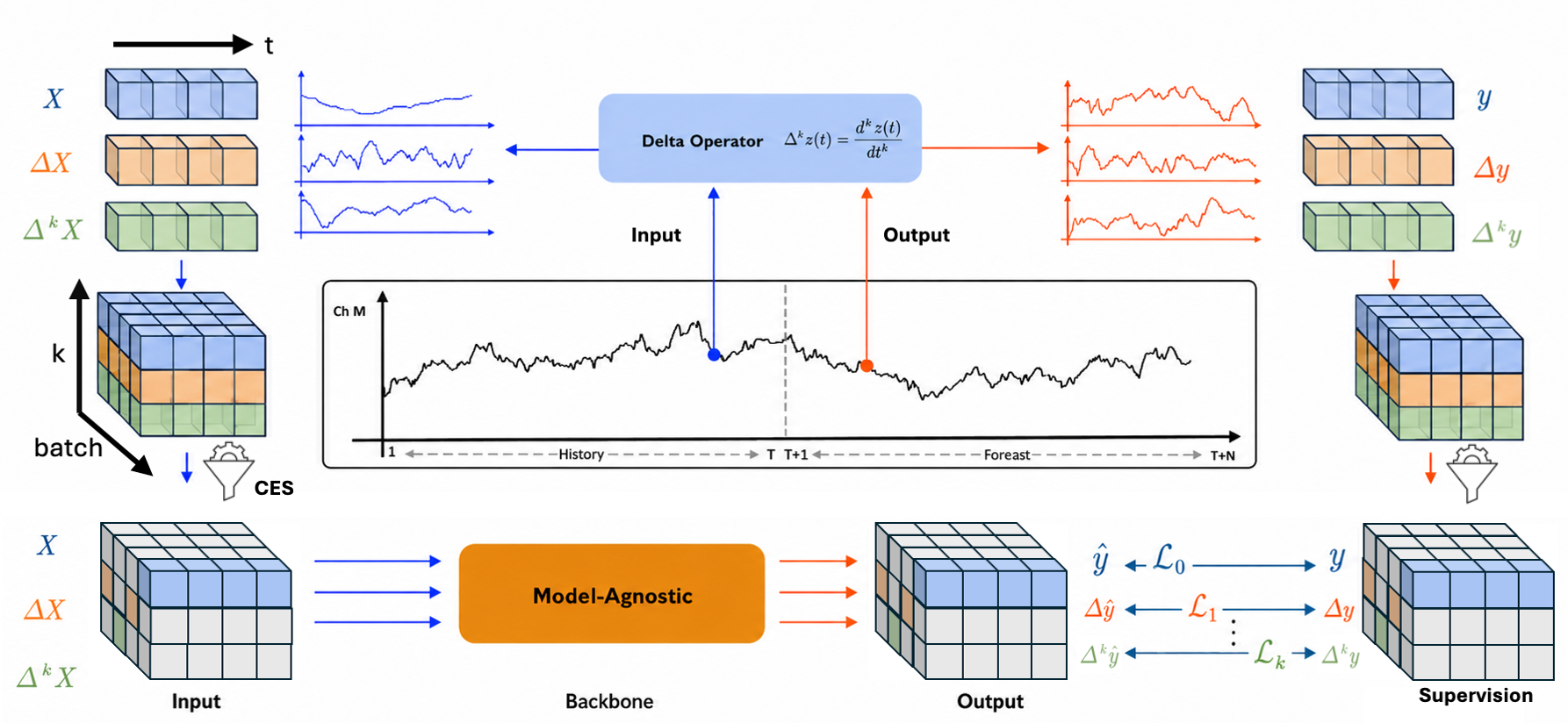}}
    \caption{Overview of our DeltaTSF, a model-agnostic training paradigm for existing forecasting backbones without modifying their architectures. DeltaTSF jointly models series and their temporal derivatives within a shared architecture, coupling their predictions across time to enforce temporally structural alignment.}

    \label{fig:main}
  
\end{figure*}

Our DeltaTSF is illustrated in Figure~\ref{fig:main}, outlining how derivative representations are learned.

\subsection{Problem Setup and Notation}
\label{sec:notation}

We consider the multivariate time-series forecasting problem.
Let $\{\mathbf{x}_t\}_{t=1}^{T}$ with $\mathbf{x}_t\in\mathbb{R}^{M}$ denote an observed history window of length $T$,
and let $\{\mathbf{y}_t\}_{t=T+1}^{T+H}$ with $\mathbf{y}_t\in\mathbb{R}^{M}$ denote the future horizon of length $H$ to be predicted.
For compactness, we stack the observations as
\begin{equation}
\mathbf{x} = [\mathbf{x}_1, \ldots, \mathbf{x}_T] \in \mathbb{R}^{T\times M},
\end{equation}
\begin{equation}
\mathbf{y} = [\mathbf{y}_{T+1}, \ldots, \mathbf{y}_{T+H}] \in \mathbb{R}^{H\times M}.
\end{equation}

The goal is to learn a forecasting function $f_{\theta}$
\begin{equation}
\mathbf{y} = f_{\theta}(\mathbf{x}):\mathbb{R}^{T\times M}\rightarrow \mathbb{R}^{H\times M},
\label{eq:backbone}
\end{equation}
which maps the historical observations $\mathbf{x}_{1:T}$ to future predictions $\mathbf{y}_{T+1:T+H}$.
Here, $\theta$ denotes the learnable parameters of a deep forecasting model. The backbone $f_{\theta}$ can be instantiated by any off-the-shelf forecaster,
such as Transformer-based PatchTST or MLP-based architectures.

\subsection{Temporal \& Dynamical Structure of Time Series.}
Multivariate time series can be viewed as discrete observations of an underlying continuous-time dynamical process.
Let $\mathbf{z}(t)\in\mathbb{R}^{M}$ denote a continuous-time signal evolving over $t\in\mathbb{R}$.
In practice, we observe $\mathbf{z}(t)$ at uniformly sampled time points $t_k = k\Delta t$,
yielding a discrete sequence $\mathbf{z}_k \in \mathbb{R}^{M}$.
In the forecasting setting, the observed history corresponds to
$\mathbf{x}_t=\mathbf{z}_t$ for $t=1,\ldots,T$,
while the future horizon corresponds to
$\mathbf{y}_t=\mathbf{z}_t$ for $t=T+1,\ldots,T+H$.

The temporal dynamics of $\mathbf{z}(t)$ are characterized by its time derivatives.
The first-order derivative
\begin{equation}
\dot{\mathbf{z}}(t)=\frac{d\mathbf{z}(t)}{dt}
\end{equation}
captures instantaneous velocity or trend,
while the second-order derivative
\begin{equation}
\ddot{\mathbf{z}}(t)=\frac{d^{2}\mathbf{z}(t)}{dt^{2}}
\end{equation}
encodes acceleration and curvature, providing a compact description of the underlying dynamics.

Since only discrete samples are available, these continuous-time derivatives must be approximated numerically.
Using finite differences with sampling interval $\Delta t$, the first-order derivative at time index $t$ is approximated by
\begin{equation}
\dot{\mathbf{z}}(t) \approx \frac{\mathbf{z}_{t}-\mathbf{z}_{t-1}}{\Delta t},
\qquad t\ge2,
\label{eq:fd_first_discrete}
\end{equation}
and the second-order derivative by
\begin{equation}
\ddot{\mathbf{z}}(t) \approx
\frac{\mathbf{z}_{t}-2\mathbf{z}_{t-1}+\mathbf{z}_{t-2}}{(\Delta t)^2},
\qquad t\ge3.
\label{eq:fd_second_discrete}
\end{equation}
These approximations are applied element-wise across the $M$ variables and hold for both
$\mathbf{x}_{1:T}$ and $\mathbf{y}_{T+1:T+H}$.

More generally, the $k$-th order forward finite difference of a sequence
$\mathbf{v}=[\mathbf{v}_1,\ldots,\mathbf{v}_L]\in\mathbb{R}^{L\times M}$ can be
expressed as a linear operator:
\begin{equation}
\mathbf{v}^{(k)}=\Delta^{k}\mathbf{v} = \mathbf{D}_{k} \mathbf{v},
\end{equation}
\begin{equation}
\mathbf{D}_{k} \in \mathbb{R}^{(L-k)\times L},
\quad k = 1,2,\dots,L-1,
\label{eq:delta_k}
\end{equation}
where $\mathbf{D}_{k}$ is the $k$-th order difference matrix constructed recursively as
\begin{equation}
\mathbf{D}_{1} = 
\begin{bmatrix}
-1 & 1 & 0 & \cdots & 0 \\
0 & -1 & 1 & \cdots & 0 \\
\vdots & & \ddots & \ddots & \vdots \\
0 & \cdots & 0 & -1 & 1
\end{bmatrix},
\qquad
\mathbf{D}_{k} = \mathbf{D}_{1} \mathbf{D}_{k-1}.
\end{equation}
This formulation provides a compact, linear-algebraic representation of first-, second-, and higher-order
temporal differences. To maintain consistent input lengths across all orders, 
we apply padding (i.e., postpending zeros) so that 
${\mathbf{x}}^{(k)} \in \mathbb{R}^{T\times M}$ for all $k$, enabling them to be
processed by the same backbone $f_{\theta}$ without architectural modifications.

\subsection{Forecasting as Trajectory Consistency}
Under this formulation, forecasting the future horizon $\mathbf{y}_{T+1:T+H}$ can be interpreted as enforcing consistency of a latent continuous trajectory $\mathbf{z}(t)$ beyond the last observed time point $t_T$. A local Taylor expansion around $t_T$ gives
\begin{equation}
\mathbf{z}(t_T+\tau)
\approx
\mathbf{z}(t_T)
+ \tau \dot{\mathbf{z}}(t_T)
+ \frac{\tau^2}{2}\ddot{\mathbf{z}}(t_T)
+ \mathcal{O}(\tau^3),
\label{eq:taylor_cont}
\end{equation}
which shows that accurate long-horizon prediction depends not only on the current state
$\mathbf{x}_T$ but also on its temporal derivatives.

In discrete time, the derivative terms in Eq.~\eqref{eq:taylor_cont} are approximated using finite differences
computed from $\mathbf{x}_{1:T}$ and $\mathbf{y}_{T+1:T+H}$.
These discrete operators act as practical surrogates for velocity and acceleration,
enabling derivative-aware supervision while remaining fully data-driven.

\paragraph{Training loss.}
The overall objective is defined as a weighted sum of mean-squared errors across multiple derivative orders:
\begin{equation}
\mathcal{L}(\theta)
=
\sum_{k=0}^{K}\mathcal{L}_k(\theta)
=
\sum_{k=0}^{K}
\lambda_{k}\,
\mathbb{E}_{(\mathbf{x},\mathbf{y})\sim \mathcal{D}}
\left[
\left\|
f_{\theta}\!\left({\mathbf{x}}^{(k)}\right)
-
\mathbf{y}^{(k)}
\right\|_{2}^{2}
\right],
\label{eq:final_loss}
\end{equation}
where $\|\cdot\|_{2}^{2}$ denotes the squared Frobenius norm over the prediction horizon and variables.

The case $k=0$ corresponds to the standard mean-squared error on future values.
Setting $K=1$ augments the objective with an auxiliary loss on the first-order temporal differences,
thereby enforcing \emph{gradient coupling} between predictions and targets.
Further setting $K=2$ additionally introduces a second-order difference loss,
which imposes \emph{curvature coupling} and encourages consistency in temporal acceleration.

\subsection{Shared parameterization $f_{\theta}$ across temporal derivative orders.}
Although {$\mathbf{x}$, $\Delta\mathbf{x}$, $\Delta^{2}\mathbf{x}$, ...} emphasize different temporal derivative components,
they are all derived from the same underlying latent process $\mathbf{x}(t)$ sampled over time.
Jointly modeling these quantities with a shared backbone implicitly constrains the learned representation
to be consistent with a single dynamical system.
From an optimization perspective, this induces gradient coupling across multiple temporal orders,
biasing the solution toward predictors whose outputs agree simultaneously
at the level of values (state), first-order changes (trend), and second-order changes (curvature).
Such coupling serves as a structural regularizer that promotes temporal coherence and dynamical consistency.

Importantly, this joint training also encourages an approximate \emph{commutativity} 
between the forecasting operator and temporal differencing across multiple orders.
Ideally, for a consistent predictor $f_{\theta}$, applying a $k$-th order finite difference 
after forecasting should yield results similar to forecasting on $k$-th order differenced inputs, i.e.,
\begin{equation}
\Delta^{k} \big( f_{\theta}(\mathbf{x}) \big)
\;\approx\;
f_{\theta}(\Delta^{k} \mathbf{x}),
\qquad k = 1, 2, \dots, K,
\label{eq:commutativity_k}
\end{equation}
where $K$ is the highest order of differences included in training.
By enforcing supervision on both sides of Eq.~\eqref{eq:commutativity_k} through shared parameters,
the model is encouraged to learn representations for which prediction and temporal differencing 
approximately commute at multiple temporal scales.
This reduces inconsistency across orders, mitigates error accumulation in multi-step forecasting,
and improves long-horizon stability in practice. We show in Appendix~\ref{app:taylorconsistency} our training framework implicitly induces Taylor Consistency. 

\subsection{Compute-Equivalent Sampling (CES)}
To address the increasing training cost that scales with $K$, we introduce a \textit{Compute-Equivalent Sampling (CES)} strategy. Instead of concatenating $K$ additional derivative series for each training sample, each series is assigned a probability of $1/(K+1)$ to be replaced by one of its first $K$-th order derivatives, or to remain unchanged. This stochastic selection preserves the overall batch size and reduces the computational overhead by a factor of $1/(K+1)$, resulting in training complexity comparable to baseline methods, as illustrated in Figure~\ref{fig:main}. Empirically, ablating CES from DeltaTSF leads to only negligible performance degradation, demonstrating that CES achieves substantial computational savings while largely preserving the effectiveness, shown in Table ~\ref{tab:deltaTSF_cost}.

\section{Experiments}
\label{sec:experiments}

\subsection{Experimental Setup}
\label{sec:exp_setup}

\paragraph{Datasets.}
We conduct experiments on seven real-world multivariate time series benchmarks, including ETT (ETTh1, ETTh2, ETTm1, ETTm2) \cite{zhou2021informer}, Weather \cite{zhou2021informer}, and Electricity (ECL) \cite{li2019enhancing}.
Following standard practice in long-term forecasting, we adopt the official data preprocessing procedures and train/validation/test splits provided by prior works \cite{nie2023timeseriesworth64,liu2023itransformer}.
Detailed dataset descriptions are provided in Appendix~\ref{app:dataset}.

\paragraph{Baselines}
To evaluate the effectiveness and generality of the proposed method, we compare DeltaTSF with a diverse set of representative state-of-the-art time series forecasting baselines spanning different modeling paradigms. Specifically, our comparison includes Transformer-based models (\textit{PatchTST} \cite{nie2023timeseriesworth64}, \textit{iTransformer} \cite{liu2023itransformer}, and \textit{PhaseFormer} \cite{khan2025phaseformer}), backbone-substitution methods based on linear, CNN, and MLP-style designs (\textit{DLinear} \cite{zeng2023transformers}, \textit{TimesNet} \cite{wu2022timesnet}, \textit{TimeMixer} \cite{wang2024timemixer}, \textit{PatchMLP} \cite{tang2025unlocking}, and \textit{TimeBase} \cite{huangtimebase}), as well as a model-agnostic training-loss baseline (\textit{Patch-wise Loss} \cite{kudrat2025patch}). Together, these baselines cover diverse design principles, including patch-wise temporal modeling, variable-wise attention, decomposition-based forecasting, periodic/spectral modeling, multiscale mixing, temporal basis representations, and output-side structural supervision. All baseline results are obtained using official implementations or original settings whenever available, with careful tuning to ensure fair comparison. Detailed descriptions of the baseline models are provided in Appendix~\ref{appendix:baseline}.

\paragraph{Implementation Details.}
We use the official implementations of all baselines and keep their backbone architectures unchanged, so the model size remains identical to the original backbones. When integrating the proposed method, we only tune the derivative loss weight, with minor optimization adjustments applied only when necessary for stable training. All experiments are implemented in PyTorch and conducted on NVIDIA RTX 4090 GPUs. Additional details are provided in Appendix~\ref{app:implementation}.

\subsection{Main Results}
\label{sec:main_results}

\begin{table*}[t]
\centering
\scriptsize
\setlength{\tabcolsep}{4pt}
\renewcommand{\arraystretch}{1.08}
\caption{Average long-term multivariate forecasting accuracies on six datasets with the input length set to 720. Our method DeltaTSF is applied to various forecating backbones. Full results are provided in Appendix Table~\ref{tab:long_term_results_no_horizon}.}
\label{tab:avg_long_term_results}
\resizebox{\textwidth}{!}{
\begin{tabular}{lcccccccccccccccc}
\toprule
\multirow{3}{*}{Dataset}
& \multicolumn{4}{c}{PatchTST}
& \multicolumn{4}{c}{iTransformer}
& \multicolumn{4}{c}{TimeMixer}
& \multicolumn{4}{c}{TimesNet} \\
\cmidrule(lr){2-5}\cmidrule(lr){6-9}\cmidrule(lr){10-13}\cmidrule(lr){14-17}
& \multicolumn{2}{c}{ori} & \multicolumn{2}{c}{+delta}
& \multicolumn{2}{c}{ori} & \multicolumn{2}{c}{+delta}
& \multicolumn{2}{c}{ori} & \multicolumn{2}{c}{+delta}
& \multicolumn{2}{c}{ori} & \multicolumn{2}{c}{+delta} \\
\cmidrule(lr){2-3}\cmidrule(lr){4-5}
\cmidrule(lr){6-7}\cmidrule(lr){8-9}
\cmidrule(lr){10-11}\cmidrule(lr){12-13}
\cmidrule(lr){14-15}\cmidrule(lr){16-17}
& MSE & MAE & MSE & MAE & MSE & MAE & MSE & MAE & MSE & MAE & MSE & MAE & MSE & MAE & MSE & MAE \\
\midrule
ETTh1       & 0.420 & 0.440 & \textbf{0.396} & \textbf{0.421} & 0.454 & 0.467 & \textbf{0.423} & \textbf{0.446} & 0.454 & 0.474 & \textbf{0.421} & \textbf{0.440} & 0.504 & 0.499 & \textbf{0.444} & \textbf{0.469} \\
ETTh2       & 0.344 & 0.391 & \textbf{0.321} & \textbf{0.376} & 0.392 & 0.422 & \textbf{0.373} & \textbf{0.409} & 0.379 & 0.426 & \textbf{0.349} & \textbf{0.395} & 0.433 & 0.455 & \textbf{0.396} & \textbf{0.414} \\
ETTm1       & 0.355 & 0.385 & \textbf{0.345} & \textbf{0.383} & 0.371 & 0.401 & \textbf{0.367} & \textbf{0.398} & 0.381 & 0.414 & \textbf{0.360} & \textbf{0.387} & 0.408 & 0.425 & \textbf{0.372} & \textbf{0.396} \\
ETTm2       & 0.251 & 0.319 & \textbf{0.250} & \textbf{0.314} & 0.279 & 0.338 & \textbf{0.268} & \textbf{0.328} & 0.283 & 0.347 & \textbf{0.274} & \textbf{0.320} & 0.300 & 0.354 & \textbf{0.275} & \textbf{0.332} \\
Weather     & 0.224 & 0.264 & \textbf{0.221} & \textbf{0.262} & 0.233 & 0.273 & {0.242} & {0.278} & 0.238 & 0.281 & \textbf{0.234} & \textbf{0.275} & 0.254 & 0.293 & \textbf{0.244} & \textbf{0.285} \\
Electricity & 0.169 & 0.266 & \textbf{0.158} & \textbf{0.253} & 0.166 & 0.264 & \textbf{0.160} & \textbf{0.257} & 0.168 & 0.264 & \textbf{0.161} & \textbf{0.255} & 0.238 & 0.334 & \textbf{0.206} & \textbf{0.305} \\
\bottomrule
\end{tabular}
}
\end{table*}

As shown in Table~\ref{tab:avg_long_term_results}, DeltaTSF consistently improves forecasting performance across multiple benchmark datasets and backbone models. The gains are observed across heterogeneous model families, including Transformer-based and MLP-style architectures, indicating that the proposed derivative-based dynamic coherence acts as a model-agnostic enhancement. These results support the effectiveness of DeltaTSF as a simple yet effective training paradigm that complements existing forecasting models.

In Table~\ref{tab:avg_results_updated}, we present the average performance of our proposed DeltaTSF (built on the PatchTST backbone) and compare it with several strong baselines across multiple forecasting horizons. Our method enables a standard backbone such as PatchTST to achieve superior forecasting accuracy compared to recent specialized architectures, including PhaseFormer, attributed to the proposed dynamic-aware training process that explicitly enforces consistent derivative representations. This highlights the effectiveness of DeltaTSF in enhancing existing models without architectural modifications. Additional experiments and analyses reported in Appendix \ref{app:additional-result}.

\begin{table*}[t]
\centering
\small
\setlength{\tabcolsep}{4pt}
\renewcommand{\arraystretch}{1.08}
\caption{Average long-term multivariate forecasting results on six datasets.Here the input sequence length fixed to 720 for all methods. In each row, the best and second-best results for each metric are highlighted in \textcolor{red}{red} and \textcolor{blue}{blue}, respectively. Full results are provided in Appendix Table ~\ref{tab:full_results_updated}}
\label{tab:avg_results_updated}
\resizebox{\textwidth}{!}{
\begin{tabular}{lcccccccccccccc}
\toprule
Dataset
& \multicolumn{2}{c}{DeltaTSF}
& \multicolumn{2}{c}{PhaseFormer}
& \multicolumn{2}{c}{PatchTST}
& \multicolumn{2}{c}{iTransformer}
& \multicolumn{2}{c}{TimeBase}
& \multicolumn{2}{c}{PatchMLP}
& \multicolumn{2}{c}{DLinear} \\
\cmidrule(lr){2-3}\cmidrule(lr){4-5}\cmidrule(lr){6-7}\cmidrule(lr){8-9}\cmidrule(lr){10-11}\cmidrule(lr){12-13}\cmidrule(lr){14-15}
& MSE & MAE & MSE & MAE & MSE & MAE & MSE & MAE & MSE & MAE & MSE & MAE & MSE & MAE \\
\midrule
ETTh1       & \textcolor{red}{\textbf{0.396}} & 0.421 & \textcolor{blue}{\textbf{0.403}} & \textcolor{red}{\textbf{0.415}} & 0.420 & 0.440 & 0.454 & 0.467 & 0.404 & \textcolor{blue}{\textbf{0.416}} & 0.483 & 0.485 & 0.437 & 0.448 \\
ETTh2       & \textcolor{red}{\textbf{0.321}} & \textcolor{red}{\textbf{0.376}} & 0.347 & \textcolor{blue}{\textbf{0.389}} & \textcolor{blue}{\textbf{0.344}} & 0.391 & 0.392 & 0.422 & 0.356 & 0.401 & 0.392 & 0.427 & 0.481 & 0.471 \\
ETTm1       & \textcolor{red}{\textbf{0.345}} & 0.383 & \textcolor{blue}{\textbf{0.347}} & \textcolor{red}{\textbf{0.374}} & 0.355 & 0.385 & 0.371 & 0.401 & 0.357 & \textcolor{blue}{\textbf{0.381}} & 0.372 & 0.396 & 0.361 & 0.383 \\
ETTm2       & \textcolor{red}{\textbf{0.250}} & \textcolor{red}{\textbf{0.314}} & \textcolor{blue}{\textbf{0.251}} & \textcolor{blue}{\textbf{0.314}} & 0.251 & 0.319 & 0.279 & 0.338 & 0.251 & 0.314 & 0.279 & 0.333 & 0.271 & 0.337 \\
Weather     & \textcolor{red}{\textbf{0.221}} & \textcolor{blue}{\textbf{0.262}} & \textcolor{blue}{\textbf{0.223}} & \textcolor{red}{\textbf{0.261}} & 0.224 & 0.264 & 0.233 & 0.273 & 0.227 & 0.263 & 0.230 & 0.270 & 0.243 & 0.299 \\
Electricity & \textcolor{red}{\textbf{0.158}} & \textcolor{blue}{\textbf{0.253}} & \textcolor{blue}{\textbf{0.161}} & \textcolor{red}{\textbf{0.250}} & 0.169 & 0.266 & 0.166 & 0.264 & 0.167 & 0.258 & 0.178 & 0.280 & 0.169 & 0.272 \\
\bottomrule
\end{tabular}
}
\end{table*}

\subsection{Ablation Study}\label{sec:abl}

\begin{figure}
    \centering
    \includegraphics[width=1.0\linewidth]{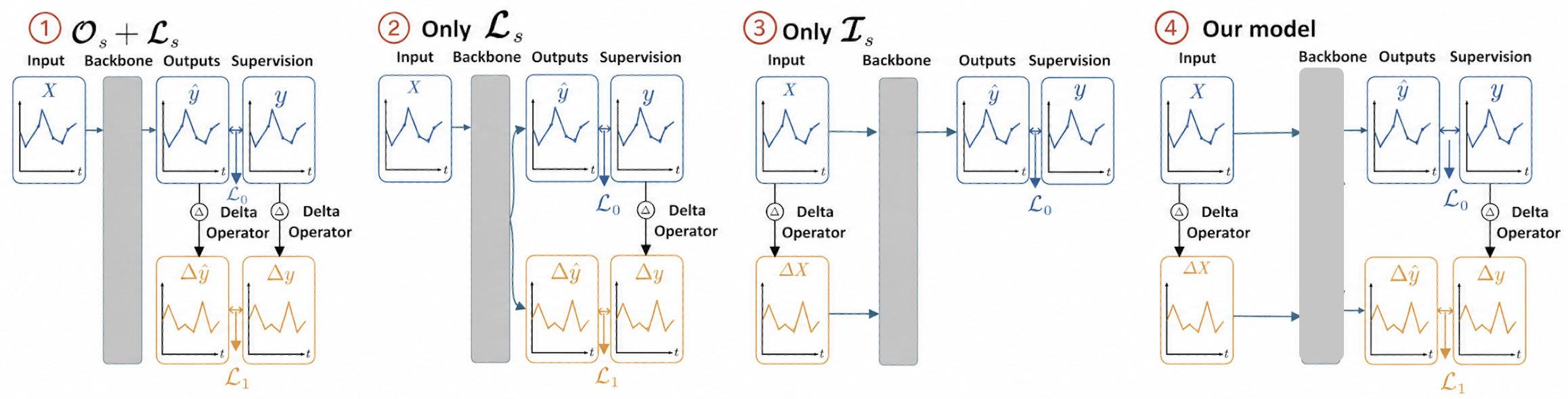}
    \caption{Illustration of the ablation studies in Section~\ref{sec:abl}, showing various approach to incorporate derivative representations into the forecasting backbone.}
    \label{fig:derivativechoice}
\end{figure}
To gain deeper insight into the performance of DeltaTSF, we conduct comprehensive ablation studies to disentangle the effects of individual design components, with a particular focus on how temporal derivative information is integrated into the forecasting pipeline. These experiments are designed to identify which components are essential for effective and stable learning of temporal dynamics.

\subsubsection{When and Where to Model Temporal Derivatives}

To answer the questions, we study the impact of incorporating derivative representations into the forecasting backbone in various ways, by introducing temporal derivative information at different stages, including the input, output, loss function, and their combinations, shown in Figure~\ref{fig:derivativechoice}.

To empirically validate these design choices, we construct a set of controlled variants that systematically introduce temporal derivative representations at different stages of the forecasting pipeline, including the input stage \textbf{(I$_S$)}, output stage \textbf{(O$_S$)}, and auxiliary loss stage \textbf{(L$_S$)}, as well as their combinations. All variants share the same backbone architecture, optimization strategy, and training protocol, ensuring that any performance differences can be attributed solely to how and where derivative information is incorporated.

\paragraph{Output-stage Derivative Supervision with Auxiliary Loss (O$_S$+L$_S$).}
This variant introduces explicit prediction of both future values and derivatives at the output stage, with supervision applied to both via an auxiliary consistency loss. While this enforces derivative alignment at the prediction level, it does not incorporate derivative information into the input representation. As shown in Table~\ref{tab:ablation_components}, this design improves over the baseline but remains inferior DeltaTSF, suggesting that modeling derivatives only at the output stage is insufficient for capturing coherent temporal dynamics.

\paragraph{Auxiliary Loss Only (L$_S$).}
In this setting, temporal derivatives are computed post hoc from predicted values and used solely as an auxiliary regularization term. This corresponds to enforcing derivative consistency only at the objective level, without affecting either representation learning or prediction structure. The marginal improvements reported in Table~\ref{tab:ablation_components} indicate that post hoc constraints alone provide weak supervision for learning temporally coherent representations.

\paragraph{Input-stage Derivative Incorporation Only (I$_S$).}
This variant augments the input sequence with derivative features while predicting only future values. Although this provides additional local dynamical context at the representation level, it does not explicitly constrain higher-order temporal structure in the prediction space. As shown in Table~\ref{tab:ablation_components}, input-only incorporation is insufficient to consistently recover the benefits of full derivative modeling, highlighting the limitation of restricting temporal dynamics to the input stage.

\paragraph{Joint Input, Output, and Loss (I$_S$+O$_S$+L$_S$).}
DeltaTSF integrates temporal derivatives across all stages of the pipeline, jointly grounding input representations, constraining outputs, and enforcing consistency through auxiliary objectives. This design effectively establishes a multi-stage discrete temporal consistency mechanism over the forecast horizon. As shown in Table~\ref{tab:ablation_components}, this unified formulation consistently achieves the best performance across datasets, demonstrating that coherent temporal dynamics emerge only when derivative information is incorporated holistically across all stages rather than in isolation.

\begin{table*}[t]
\centering

\begin{subtable}[t]{0.48\textwidth}
\centering
\scriptsize
\caption{Different ways of incorporating derivatives. Full results are presented in Appendix Table~\ref{tab:ablationmixupfull}.}
\label{tab:ablation_components}
\setlength{\tabcolsep}{5pt}
\begin{tabular}{lcccc}
\toprule
Setting & \multicolumn{2}{c}{ETTh1-AVG} & \multicolumn{2}{c}{ETTh2-AVG} \\
\cmidrule(lr){2-3}\cmidrule(lr){4-5}
 & MSE & MAE & MSE & MAE \\
\midrule
$\mathcal{O}_S{+}\mathcal{L}_S$ & 0.422 & 0.441 & 0.332 & 0.382 \\
Only $\mathcal{L}_S$            & 0.408 & 0.431 & 0.327 & 0.381 \\
Only $\mathcal{I}_S$            & 0.411 & 0.433 & 0.332 & 0.384 \\
Ours $(\mathcal{I}_S{+}\mathcal{O}_S{+}\mathcal{L}_S)$ & \textbf{0.396} & \textbf{0.421} & \textbf{0.322} & \textbf{0.375} \\
\bottomrule
\end{tabular}
\end{subtable}
\hfill
\begin{subtable}[t]{0.40\textwidth}
\centering
\small
\caption{Ablation of CES (MSE / MAE).}
\label{tab:deltaTSF_cost}
\setlength{\tabcolsep}{6pt}
\begin{tabular}{lcc}
\toprule
Dataset & DeltaTSF & wo CES \\
\midrule
ETTh1   & 0.396 / 0.421 & 0.396 / 0.421 \\
ETTh2   & 0.321 / 0.376 & 0.322 / 0.375 \\
ETTm2   & 0.250 / 0.314 & 0.248 / 0.310 \\
Weather & 0.221 / 0.262 & 0.219 / 0.261 \\
\bottomrule
\end{tabular}
\end{subtable}

\caption{Ablation studies on derivative integration and CES.}
\label{tab:ablation_main}
\end{table*}

\subsubsection{Impact of CES}
The CES strategy ensures the same training cost as the original time-series backbone. As shown in Table~\ref{tab:deltaTSF_cost}, CES achieves performance comparable to the full-data setting, with a negligible performance degradation on the order of $10^{-3}$, while reducing computational cost by at least 50\%.

\subsubsection{Impact of Higher Taylor Order}

\begin{table}[t]
\centering
\scriptsize
\caption{Impact of higher-order temporal differences in DeltaTSF with PatchTST as the backbone on ETTh1.}

\label{tab:patchtst_diff_order}
\setlength{\tabcolsep}{6pt}
\begin{tabular}{lccccc}
\toprule
Horizon & PatchTST (MSE/MAE) & +K=1 (MSE/MAE) & +K=2 (MSE/MAE) & +K=3 (MSE/MAE) & +K=4 (MSE/MAE) \\
\midrule
96  & 0.377 / 0.408 & 0.362 / 0.395 & 0.364 / 0.397 & 0.366 / 0.398 & 0.367 / 0.400 \\
192 & 0.413 / 0.431 & 0.396 / 0.416 & 0.396 / 0.418 & 0.397 / 0.418 & 0.398 / 0.419 \\
336 & 0.436 / 0.446 & 0.405 / 0.424 & 0.406 / 0.426 & 0.409 / 0.429 & 0.406 / 0.429 \\
720 & 0.455 / 0.475 & 0.420 / 0.448 & 0.422 / 0.450 & 0.427 / 0.453 & 0.425 / 0.453 \\
\midrule
AVG & 0.420 / 0.440 & 0.396 / 0.421 & 0.397 / 0.423 & 0.400 / 0.425 & 0.399 / 0.425 \\
\bottomrule
\end{tabular}
\end{table}

As shown in Table~\ref{tab:patchtst_diff_order}, introducing first-order temporal derivatives in DeltaTSF consistently improves forecasting performance over the PatchTST backbone across all horizons, reducing the average error from 0.420/0.440 to 0.396/0.421. 
We further extend the method by incorporating higher-order differences for $K=2,3,4$. 
In high-order finite differences, the noise components of the time series start to affect the signals, yet our method does not exhibit signs of overfitting or training divergence. This demonstrates that higher-order derivatives may introduce useful or noisy information depending on the dataset; however, they do not degrade the performance or stability of our method generally.

\section{Conclusion}

In this work, we study time series forecasting from a dynamical perspective by modeling temporal evolution through derivatives. We propose DeltaTSF, a dynamics-aware forecasting framework that jointly learns numerical values and their multi-order temporal derivatives within a unified architecture. By supervising predictions across both value and derivative spaces, DeltaTSF enforces cross-horizon temporal consistency during training, effectively constraining future trajectories to respect locally valid dynamical behavior. This training formulation implicitly induces a discrete Taylor-style regularization, coupling predictions at different horizons through their estimated temporal derivatives. As a result, the model is encouraged to produce forecasts that are not only accurate at individual time steps but also coherent with the underlying temporal dynamics of the data. Importantly, these constraints are realized entirely during training, requiring no explicit derivative computation or additional overhead at inference time. Comprehensive experiments on multiple benchmark datasets demonstrate that incorporating dynamic supervision leads to consistent improvements in forecasting accuracy, stability, and long-horizon coherence. The empirical results suggest that multi-order temporal dynamics provide complementary inductive biases beyond value-only supervision, improving the structural consistency of predicted trajectories.

\section{Limitation}\label{sec:limitation}
While this work presents a simple yet effective approach to improving existing forecasting backbones without requiring architectural modifications, it is intended to be complementary to ongoing research in architectural design. Although consistent improvements are observed across benchmarks, there remains non-trivial room for further performance gains in terms of MAE and MSE. Moreover, the results highlight the effectiveness of incorporating derivative representations for modeling temporal dynamics, suggesting that future work could explore dedicated architectural designs, particularly within Transformer-based frameworks, that more explicitly exploit such representations. While the proposed method introduces no additional training or inference cost at the model level, the data loading pipeline requires computation of derivatives, which may introduce additional preprocessing overhead. Future work may explore optimized data loader implementations or efficient parallelization strategies to further reduce this small overhead.

{
\small

\bibliography{example_paper}

}

\newpage
\appendix
\onecolumn
\section{Analysis of DeltaTSF}

\subsection{Taylor Consistency Induced by Our Training Framework}
\label{app:taylorconsistency}

Let us consider a local Taylor expansion of the future horizon at each predicted step $t$, starting from the last observed step $T$ up to the horizon $T+H-1$:
\begin{equation}
\mathbf{y}_{t+\tau} \approx
\mathbf{y}_t
+ \tau \, \dot{\mathbf{y}}_t
+ \frac{\tau^2}{2} \, \ddot{\mathbf{y}}_t
+ \dots
+ \frac{\tau^K}{K!} \, \mathbf{y}_t^{(K)},
\qquad
t = T, \dots, T+H-1, \;\;
\tau = 1, \dots, T+H - t,
\label{eq:taylor_every_timestep}
\end{equation}
where $\mathbf{y}_t^{(k)}$ denotes the $k$-th order discrete derivative at step $t$. 

Our multi-order derivative loss
\begin{equation}
\mathcal{L}(\theta)
=
\sum_{k=0}^{K} \lambda_k \, 
\mathbb{E}_{(\mathbf{x},\mathbf{y})\sim \mathcal{D}}
\Big[
\big\| f_\theta({\mathbf{x}}^{(k)}) - \mathbf{y}^{(k)} \big\|_2^2
\Big]
\label{eq:multiorder_loss_repeat2}
\end{equation}
enforces that the predictor $f_\theta$ matches both the values ($k=0$) and the first-, second-, ..., $K$-th order discrete derivatives ($k=1,\dots,K$) of the target sequence. By jointly training on multiple derivative orders, our framework enforces an approximate commutativity:
\begin{equation}
\Delta^k \big( f_\theta(\mathbf{x}) \big) \approx f_\theta(\Delta^k \mathbf{x}), 
\qquad k = 1,\dots,K,
\end{equation}
which ensures that the derivatives of predicted sequences are consistent with the derivatives computed from the predictions themselves.

\paragraph{Taylor consistency.} 
Substituting finite-difference approximations for $\mathbf{y}_t^{(k)}$ and ignoring higher-order terms, we have
\begin{equation}
f_\theta(\mathbf{x})_t \approx \mathbf{x}_t
+ \Delta t \, (\Delta \mathbf{y})_t
+ \frac{(\Delta t)^2}{2} \, (\Delta^2 \mathbf{y})_t
+ \dots
+ \frac{(\Delta t)^K}{K!} \, (\Delta^K \mathbf{y})_t,
\qquad t = T, \dots, T+H-1.
\end{equation}
This corresponds directly to a truncated discrete-time Taylor expansion.

Thus, our training framework guarantees that the predictions are consistent not only in value but also in trend, curvature, and higher-order temporal changes at every predicted timestep:
\begin{equation}
f_\theta(\mathbf{x})_{t+\tau} \;\approx\; 
\mathbf{y}_t + \sum_{k=1}^{K} \frac{\tau^k}{k!} \, (\Delta^k \mathbf{y})_t,
\qquad t = T, \dots, T+H-1, \;\;
\tau = 1, \dots, T+H - t.
\end{equation}

We refer to this property as \emph{Taylor consistency}: the model respects the underlying temporal dynamics of the latent trajectory across multiple derivative orders, yielding predictions that are dynamically coherent and stable over long horizons.

\newpage

\section{Baseline Models}
\label{appendix:baseline}
We adopt a representative set of recent time-series forecasting methods that cover diverse representation spaces and intervention dimensions, including Transformer backbone redesign, non-Transformer backbone substitution, and model-agnostic training-paradigm improvement. Specifically, our comparison set includes Transformer-based models (\textbf{PatchTST}, \textbf{iTransformer}, and \textbf{PhaseFormer}), backbone-substitution methods based on linear, CNN, and MLP-style architectures (\textbf{DLinear}, \textbf{TimesNet}, \textbf{TimeMixer}, \textbf{PatchMLP}, and \textbf{TimeBase}), as well as a model-agnostic training-loss method (\textbf{Patch-wise Loss}). This selection enables a structured comparison between architectural innovation and training-strategy improvement in long-term time-series forecasting.

\begin{itemize}
    \item \textbf{PatchTST} \cite{nie2023timeseriesworth64}: It segments the input time series into temporal patches and applies a Transformer along the temporal dimension, providing a strong patch-based Transformer baseline for long-horizon forecasting.

    \item \textbf{DLinear} \cite{zeng2023transformers}: It decomposes the input series into trend and seasonal components and performs forecasting using lightweight linear layers, serving as a strong decomposition-based linear baseline.

    \item \textbf{TimesNet} \cite{wu2022timesnet}: It reshapes one-dimensional time series into two-dimensional representations across multiple periods and employs convolutional operations to capture both intra-period and inter-period temporal dependencies.

    \item \textbf{TimeMixer} \cite{wang2024timemixer}: It decomposes time series into multiple temporal scales and mixes the resulting multi-scale representations through MLP-style operations, offering an efficient decomposition-based alternative to Transformer backbones.

    \item \textbf{iTransformer} \cite{liu2023itransformer}: It inverts the standard tokenization strategy by treating variates as tokens and applies Transformer layers along the channel dimension to explicitly model inter-variable dependencies.

    \item \textbf{PatchMLP} \cite{tang2025unlocking}: It replaces the Transformer backbone in patch-based forecasting with an MLP architecture, highlighting the effectiveness of patch representations beyond self-attention.

    \item \textbf{TimeBase} \cite{huangtimebase}: It extracts a small set of core temporal basis components and performs segment-level forecasting using an ultra-lightweight linear design, achieving high efficiency with substantially reduced model complexity.

    \item \textbf{Patch-wise Structural Loss (PSloss)} \cite{kudrat2025patch}: Unlike the backbone models above, Patch-wise Loss is a model-agnostic training objective rather than a standalone forecasting architecture. It partitions the prediction and target series into adaptive patches and introduces patch-level structural supervision by jointly matching correlation, variance, and mean, in addition to conventional point-wise loss.

    \item \textbf{PhaseFormer} \cite{khan2025phaseformer}: It introduces a phase-based representation for time-series forecasting and redesigns the Transformer architecture to better capture periodic and phase-aware temporal structures.
\end{itemize}

\section{Experiment Details}
In this section, we provide detailed descriptions of the datasets, implementation details, and experimental environments to ensure the reproducibility of our results.
\subsection{Datasets Description}\label{app:dataset}
We extensively evaluate the proposed DeltaTSF on six widely used real-world benchmarks for long-term time series forecasting, covering energy, traffic, and weather domains. These include ETTh1, ETTh2, ETTm1, ETTm2, Weather, and Electricity.

The detailed descriptions of each dataset are as follows:
\begin{itemize}
    \item \textbf{Weather} \cite{zhou2021informer}:  
    This dataset contains local climatological data recorded every 10 minutes for the year 2020 in Germany, consisting of 21 meteorological indicators.

    \item \textbf{ECL (Electricity)} \cite{li2019enhancing}:  
    It consists of hourly electricity consumption data from 321 clients between 2012 and 2014.

    \item \textbf{ETT (Electricity Transformer Temperature)} \cite{zhou2021informer}:  
    The ETT dataset collects data from electricity transformers, including two hourly-level datasets (ETTh1, ETTh2) and two 15-minute-level datasets (ETTm1, ETTm2). Each dataset contains 7 distinct variables, including oil temperature and power load features.
\end{itemize}
\paragraph{Data Preprocessing and Splitting}
We strictly follow the standard protocol utilized in PatchTST \cite{nie2023timeseriesworth64}.
Splitting: All datasets are split into training, validation, and testing sets in chronological order. The split ratio is set to 6:2:2 for the ETT datasets and 7:1:2 for the Weather, Electricity, and Traffic datasets. To handle non-stationarity and distribution shifts, we employ Reversible Instance Normalization (RevIN) \cite{kim2021reversible} within the model architecture, ensuring the preservation of the original data statistics during inference.

\subsection{Implementation Details}
\label{app:implementation}

\paragraph{Model Configuration.}
For all datasets, the input (look-back) window length is fixed to $L = 720$.
The forecasting horizon $H$ is selected from $\{96, 192, 336, 720\}$.
Unless otherwise specified, the same model configuration is used across all experimental settings.

\paragraph{Optimization and Hyperparameters.}
All models are optimized using the Adam optimizer with an initial learning rate of $1 \times 10^{-4}$.
The batch size is set to 32 by default and increased to 128 when permitted by GPU memory constraints.
No additional learning rate scheduling is applied unless explicitly stated.

\paragraph{Training Protocol.}
Models are trained for a maximum of 100 epochs.
To prevent overfitting, early stopping is employed based on validation loss, with training terminated if no improvement is observed for 5 consecutive epochs.
The model parameters corresponding to the best validation performance are retained for evaluation.

\textbf{Model Configuration.} 
For all datasets, the input seq-len is fixed to $L = 720$. The forecasting horizon $H$ is selected from $\{96, 192, 336, 720\}$. Unless otherwise specified, DeltaTSF uses PatchTST as its backbone throughout this paper. For Tables~9 and~10, the results of the other baseline models are directly taken from the original PhaseFormer and TimeBase papers.

\paragraph{Hardware and Software.}
All experiments are implemented in PyTorch and conducted on a workstation equipped with eight NVIDIA RTX 4090 GPUs.

\subsection{Evaluation Metrics}

We evaluate forecasting performance using \emph{Mean Squared Error (MSE)} and
\emph{Mean Absolute Error (MAE)} over the prediction horizon.
Let
\(\mathbf{y} = [\mathbf{y}_{T+1}, \ldots, \mathbf{y}_{T+H}] \in \mathbb{R}^{H \times M}\)
denote the ground-truth future values, and
\(\hat{\mathbf{y}} = [\hat{\mathbf{y}}_{T+1}, \ldots, \hat{\mathbf{y}}_{T+H}] \in \mathbb{R}^{H \times M}\)
denote the corresponding predictions produced by the forecasting model \(f_{\theta}\).

The evaluation metrics are defined as
\begin{equation}
\mathrm{MSE}
= \frac{1}{HM}
\sum_{t=T+1}^{T+H}
\left\|
\mathbf{y}_t - \hat{\mathbf{y}}_t
\right\|_2^2 ,
\end{equation}

\begin{equation}
\mathrm{MAE}
= \frac{1}{HM}
\sum_{t=T+1}^{T+H}
\left\|
\mathbf{y}_t - \hat{\mathbf{y}}_t
\right\|_1 ,
\end{equation}
where \(\|\cdot\|_2\) and \(\|\cdot\|_1\) denote the \(\ell_2\) and \(\ell_1\) norms, respectively.
Both metrics are averaged over all prediction horizons and variates.

\newpage
\section{Additional Results}\label{app:additional-result}

\subsection{Sensitivity to the Derivative Loss Weight}
\begin{figure}
    \centering
    \includegraphics[width=\linewidth]{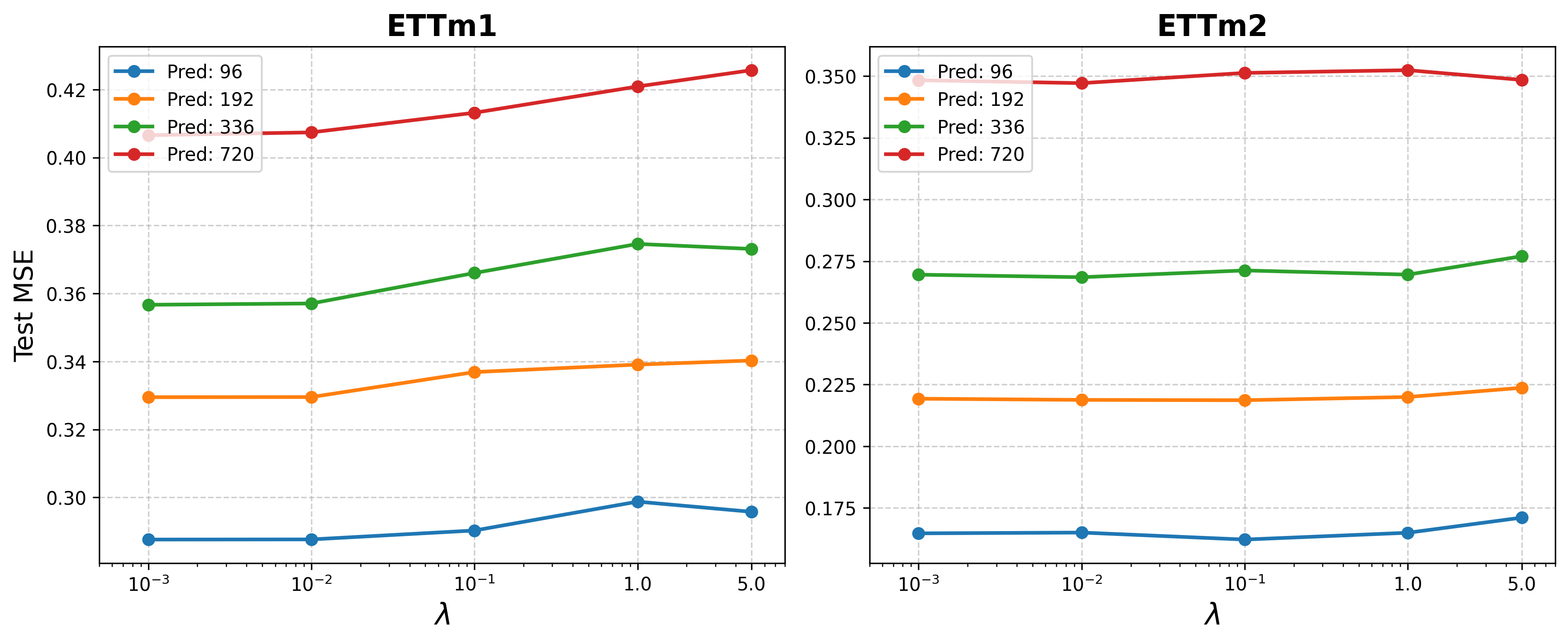}
    \caption{Sensitivity of forecasting performance to the hyperparameter $\lambda$ on the ETTm1 and ETTm2 dataset across different prediction horizons.}
    \label{fig:Hyperparameter Sensitivity}
\end{figure}

Figure~\ref{fig:Hyperparameter Sensitivity} shows the sensitivity of DeltaTSF to the derivative loss weight $\lambda$ on ETTm1 and ETTm2 under four prediction horizons. We evaluate a shared candidate set
\[
\lambda \in \{10^{-3},\,10^{-2},\,10^{-1},\,1,\,5\},
\]
and all $\lambda$ values reported in our experiments are selected from this set. In particular, for each dataset and prediction horizon, we choose $\lambda$ from the same candidate pool based on validation performance, without introducing dataset-specific search spaces. As shown in the figure, DeltaTSF is generally stable across a broad range of $\lambda$, while excessively large values may slightly degrade performance, especially on longer horizons. Overall, relatively small to moderate weights consistently provide strong results, indicating that the proposed derivative-based regularization is effective without requiring delicate hyperparameter tuning.

\subsection{Comparison with Patch-wise Structure Loss}
As shown in Table~\ref{tab:patchps_vs_deltatsf}, we compare DeltaTSF with a representative work, PSLoss~\cite{kudrat2025patch}, which improves forecasting performance without modifying the backbone architecture. We follow the same experimental setting as in~\cite{kudrat2025patch} for a fair comparison. The results show that DeltaTSF achieves competitive performance with this strong baseline, indicating that it can reach a high level among this class of enhancement methods without introducing additional training cost, while requiring less prior knowledge of spectral- or decomposition-based representations of time series compared to PSLoss.

\begin{table*}[t]
\centering
\caption{Comparison of PatchTST, PatchTST+PS, and DeltaTSF on ETTh1, ETTh2, and Weather. The results of PatchTST and PatchTST+PS are taken from the Patchwise-Loss paper, and our DeltaTSF results are obtained under the same experimental setting for a fair comparison (input length 336).}
\label{tab:patchps_vs_deltatsf}
\setlength{\tabcolsep}{4pt}
\renewcommand{\arraystretch}{1.05}
\begin{tabular}{llcccccc}
\toprule
Dataset & Horizon & \multicolumn{2}{c}{PatchTST} & \multicolumn{2}{c}{PatchTST+PS} & \multicolumn{2}{c}{DeltaTSF} \\
\cmidrule(lr){3-4} \cmidrule(lr){5-6} \cmidrule(lr){7-8}
& & MSE & MAE & MSE & MAE & MSE & MAE \\
\midrule
\multirow{5}{*}{ETTh1}
& 96  & 0.375 & 0.400 & \textbf{0.368} & 0.394 & 0.369 & \textbf{0.393} \\
& 192 & 0.414 & 0.421 & 0.407 & 0.415 & \textbf{0.405} & \textbf{0.413} \\
& 336 & 0.432 & 0.436 & 0.420 & 0.426 & \textbf{0.415} & \textbf{0.422} \\
& 720 & 0.450 & 0.466 & 0.425 & 0.449 & \textbf{0.423} & \textbf{0.447} \\
& Avg & 0.418 & 0.431 & 0.405 & 0.421 & \textbf{0.403} & \textbf{0.419} \\
\midrule
\multirow{5}{*}{ETTh2}
& 96  & 0.274 & 0.336 & \textbf{0.272} & \textbf{0.332} & 0.274 & 0.333 \\
& 192 & 0.339 & 0.380 & \textbf{0.336} & 0.375 & 0.337 & \textbf{0.374} \\
& 336 & 0.331 & 0.380 & 0.324 & 0.377 & \textbf{0.320} & \textbf{0.374} \\
& 720 & 0.378 & 0.421 & 0.376 & 0.417 & \textbf{0.368} & \textbf{0.412} \\
& Avg & 0.331 & 0.379 & 0.327 & 0.375 & \textbf{0.325} & \textbf{0.373} \\
\midrule
\multirow{5}{*}{Weather}
& 96  & 0.151 & 0.198 & \textbf{0.149} & \textbf{0.190} & 0.150 & 0.196 \\
& 192 & 0.196 & 0.242 & \textbf{0.193} & \textbf{0.235} & 0.194 & 0.239 \\
& 336 & 0.248 & 0.282 & \textbf{0.245} & \textbf{0.276} & 0.247 & 0.280 \\
& 720 & 0.319 & 0.335 & \textbf{0.318} & \textbf{0.330} & 0.319 & 0.332 \\
& Avg & 0.228 & 0.264 & \textbf{0.226} & \textbf{0.258} & 0.228 & 0.262 \\
\bottomrule
\end{tabular}
\end{table*}

\begin{table*}[t]
  \caption{Ablation study on injecting temporal difference information at different stages (lower is better).}
  \label{tab:ablationmixupfull}
  \begin{center}
    \begin{tabular}{l c cc cc cc cc}
      \toprule
      Dataset & Length 
      & \multicolumn{2}{c}{O$_S$+L$_S$} 
      & \multicolumn{2}{c}{Only I$_S$} 
      & \multicolumn{2}{c}{Only L$_S$} 
      & \multicolumn{2}{c}{Ours (I$_S$+O$_S$+L$_S$)} \\
      \cmidrule(lr){3-4} \cmidrule(lr){5-6} \cmidrule(lr){7-8} \cmidrule(lr){9-10}
      & & MSE & MAE & MSE & MAE & MSE & MAE & MSE & MAE \\
      \midrule
      ETTh1 & 96  & 0.377 & 0.409 & 0.370 & 0.403 & 0.367 & 0.401 & \textbf{0.362} & \textbf{0.395} \\
            & 192 & 0.414 & 0.432 & 0.407 & 0.426 & 0.402 & 0.422 & \textbf{0.396} & \textbf{0.416} \\
            & 336 & 0.429 & 0.443 & 0.417 & 0.434 & 0.414 & 0.432 & \textbf{0.405} & \textbf{0.424} \\
            & 720 & 0.466 & 0.480 & 0.450 & 0.468 & 0.447 & 0.467 & \textbf{0.420} & \textbf{0.448} \\
            & AVG & 0.422 & 0.441 & 0.411 & 0.433 & 0.408 & 0.431 & \textbf{0.396} & \textbf{0.421} \\
      \midrule
      ETTh2 & 96  & 0.274 & 0.337 & 0.274 & 0.338 & 0.272 & 0.336 & \textbf{0.269} & \textbf{0.333} \\
            & 192 & 0.339 & 0.380 & 0.340 & 0.382 & 0.335 & 0.379 & \textbf{0.330} & \textbf{0.375} \\
            & 336 & 0.329 & 0.384 & 0.328 & 0.385 & 0.324 & 0.383 & \textbf{0.316} & \textbf{0.377} \\
            & 720 & 0.384 & 0.427 & 0.386 & 0.429 & 0.378 & 0.425 & \textbf{0.368} & \textbf{0.417} \\
            & AVG & 0.332 & 0.382 & 0.332 & 0.384 & 0.327 & 0.381 & \textbf{0.321} & \textbf{0.376} \\
      \midrule
      ETTm1 & 96  & 0.296 & 0.352 & 0.299 & 0.348 & 0.299 & 0.348 & \textbf{0.299} & \textbf{0.352} \\
            & 192 & 0.371 & 0.372 & 0.332 & 0.369 & 0.332 & 0.369 & \textbf{0.327} & \textbf{0.370} \\
            & 336 & 0.358 & 0.392 & 0.360 & 0.387 & 0.360 & 0.387 & \textbf{0.354} & \textbf{0.390} \\
            & 720 & 0.413 & 0.419 & 0.406 & 0.413 & 0.406 & 0.413 & \textbf{0.401} & \textbf{0.418} \\
            & AVG & 0.360 & 0.384 & 0.349 & 0.379 & 0.349 & 0.354 & \textbf{0.345} & \textbf{0.383} \\
      \bottomrule
    \end{tabular}
  \end{center}
\end{table*}

\subsection{Comparison with Data Augmentation.}
We further examine whether the performance gains of the proposed method can be attributed to data augmentation effects by comparing DeltaTSF with Mixup \cite{zhang2018mixup}, a widely adopted sample-level augmentation strategy. 
As reported in Table~\ref{tab:mixup_vs_DeltaTSFfull}, Mixup provides only limited improvements over the vanilla PatchTST across datasets and generally underperforms DeltaTSF. 
This comparison indicates that the observed gains are not merely due to increased data diversity. 
Unlike Mixup, which constructs synthetic samples via linear interpolation, DeltaTSF explicitly models temporal change patterns and incorporates them as direct prediction targets, providing structured supervision that more effectively guides representation learning.

\begin{table*}[t]
  \caption{Comparative analysis of PatchTST with +MixUp and $+Delta$. Results are reported in terms of MSE / MAE (lower is better).}
  \label{tab:mixup_vs_DeltaTSFfull}
  \begin{center}
    \begin{small}
      \begin{sc}
        \begin{tabular}{llcccccc}
          \toprule
          Dataset & Horizon 
          & \multicolumn{2}{c}{PatchTST} 
          & \multicolumn{2}{c}{+MixUp} 
          & \multicolumn{2}{c}{$+Delta$} \\
          \cmidrule(lr){3-4} \cmidrule(lr){5-6} \cmidrule(lr){7-8}
          & & MSE & MAE & MSE & MAE & MSE & MAE \\
          \midrule
          \multirow{5}{*}{ETTh1}
          & 96  & 0.377 & 0.408 & 0.375 & 0.407 & \textbf{0.362} & \textbf{0.395} \\
          & 192 & 0.413 & 0.431 & 0.414 & 0.432 & \textbf{0.396} & \textbf{0.416} \\
          & 336 & 0.436 & 0.446 & 0.426 & 0.441 & \textbf{0.405} & \textbf{0.424} \\
          & 720 & 0.455 & 0.475 & 0.448 & 0.469 & \textbf{0.420} & \textbf{0.448} \\
          & AVG & 0.420 & 0.440 & 0.416 & 0.437 & \textbf{0.396} & \textbf{0.421} \\
          \midrule
          \multirow{5}{*}{ETTh2}
          & 96  & 0.276 & 0.339 & 0.274 & 0.337 & \textbf{0.269} & \textbf{0.333} \\
          & 192 & 0.342 & 0.385 & 0.339 & 0.379 & \textbf{0.330} & \textbf{0.375} \\
          & 336 & 0.364 & 0.405 & 0.330 & 0.384 & \textbf{0.316} & \textbf{0.377} \\
          & 720 & 0.395 & 0.434 & 0.383 & 0.426 & \textbf{0.368} & \textbf{0.417} \\
          & AVG & 0.344 & 0.391 & 0.332 & 0.382 & \textbf{0.321} & \textbf{0.376} \\
          \midrule
          \multirow{5}{*}{ETTm1}
          & 96  & 0.298 & 0.352 & 0.295 & 0.347 & \textbf{0.299} & \textbf{0.352} \\
          & 192 & 0.335 & 0.373 & 0.333 & 0.370 & \textbf{0.327} & \textbf{0.370} \\
          & 336 & 0.366 & 0.394 & 0.364 & 0.390 & \textbf{0.354} & \textbf{0.390} \\
          & 720 & 0.420 & 0.421 & 0.414 & 0.418 & \textbf{0.401} & \textbf{0.418} \\
          & AVG & 0.355 & 0.385 & 0.352 & 0.381 & \textbf{0.345} & \textbf{0.383} \\
          \bottomrule
        \end{tabular}
      \end{sc}
    \end{small}
  \end{center}
  \vskip -0.1in
\end{table*}

\begin{table*}[t]
\centering
\scriptsize
\setlength{\tabcolsep}{3pt}
\renewcommand{\arraystretch}{1.08}
\caption{Long-term multivariate forecasting results. The table reports MSE and MAE for different forecasting lengths $T \in \{96,192,336,720\}$. The input sequence length is fixed to 720 for all methods. Better results under the $+ delta$ setting are highlighted in bold. The results of the baseline models are directly obtained from PhaseFormer~\cite{khan2025phaseformer} and TimeBase~\cite{huangtimebase}.}

\label{tab:long_term_results_no_horizon}
\resizebox{\textwidth}{!}{
\begin{tabular}{lcccccccccccccccc}
\toprule
\multirow{3}{*}{Dataset}
& \multicolumn{4}{c}{PatchTST}
& \multicolumn{4}{c}{iTransformer}
& \multicolumn{4}{c}{TimeMixer}
& \multicolumn{4}{c}{TimesNet} \\
\cmidrule(lr){2-5}\cmidrule(lr){6-9}\cmidrule(lr){10-13}\cmidrule(lr){14-17}
& \multicolumn{2}{c}{ori} & \multicolumn{2}{c}{$+Delta$}
& \multicolumn{2}{c}{ori} & \multicolumn{2}{c}{$+Delta$}
& \multicolumn{2}{c}{ori} & \multicolumn{2}{c}{$+Delta$}
& \multicolumn{2}{c}{ori} & \multicolumn{2}{c}{$+Delta$} \\
\cmidrule(lr){2-3}\cmidrule(lr){4-5}
\cmidrule(lr){6-7}\cmidrule(lr){8-9}
\cmidrule(lr){10-11}\cmidrule(lr){12-13}
\cmidrule(lr){14-15}\cmidrule(lr){16-17}
& MSE & MAE & MSE & MAE & MSE & MAE & MSE & MAE & MSE & MAE & MSE & MAE & MSE & MAE & MSE & MAE \\
\midrule

ETTh1-96  & 0.377 & 0.408 & \textbf{0.362} & \textbf{0.395} & 0.389 & 0.421 & \textbf{0.373} & \textbf{0.406} & 0.410 & 0.441 & \textbf{0.391} & \textbf{0.420} & 0.437 & 0.454 & \textbf{0.401} & \textbf{0.433} \\
ETTh1-192 & 0.413 & 0.431 & \textbf{0.396} & \textbf{0.416} & 0.424 & 0.446 & \textbf{0.409} & \textbf{0.431} & 0.448 & 0.465 & \textbf{0.415} & \textbf{0.431} & 0.454 & 0.469 & \textbf{0.425} & \textbf{0.453} \\
ETTh1-336 & 0.436 & 0.446 & \textbf{0.405} & \textbf{0.424} & 0.456 & 0.469 & \textbf{0.423} & \textbf{0.440} & 0.482 & 0.490 & \textbf{0.426} & \textbf{0.440} & 0.494 & 0.494 & \textbf{0.466} & \textbf{0.487} \\
ETTh1-720 & 0.455 & 0.475 & \textbf{0.420} & \textbf{0.448} & 0.545 & 0.532 & \textbf{0.486} & \textbf{0.505} & 0.475 & 0.500 & \textbf{0.450} & \textbf{0.468} & 0.632 & 0.578 & \textbf{0.485} & \textbf{0.502} \\
ETTh1-AVG & 0.420 & 0.440 & \textbf{0.396} & \textbf{0.421} & 0.454 & 0.467 & \textbf{0.423} & \textbf{0.446} & 0.454 & 0.474 & \textbf{0.421} & \textbf{0.440} & 0.504 & 0.499 & \textbf{0.444} & \textbf{0.469} \\
\midrule

ETTh2-96  & 0.276 & 0.339 & \textbf{0.269} & \textbf{0.333} & 0.305 & 0.361 & \textbf{0.285} & \textbf{0.345} & 0.315 & 0.380 & \textbf{0.297} & \textbf{0.356} & 0.349 & 0.403 & \textbf{0.315} & \textbf{0.358} \\
ETTh2-192 & 0.342 & 0.385 & \textbf{0.330} & \textbf{0.375} & 0.405 & 0.421 & \textbf{0.389} & \textbf{0.411} & 0.383 & 0.415 & \textbf{0.360} & \textbf{0.395} & 0.500 & 0.488 & \textbf{0.388} & \textbf{0.400} \\
ETTh2-336 & 0.364 & 0.405 & \textbf{0.316} & \textbf{0.377} & 0.411 & 0.436 & \textbf{0.407} & \textbf{0.432} & 0.385 & 0.438 & \textbf{0.340} & \textbf{0.394} & 0.445 & 0.465 & \textbf{0.442} & \textbf{0.446} \\
ETTh2-720 & 0.395 & 0.434 & \textbf{0.368} & \textbf{0.417} & 0.448 & 0.470 & \textbf{0.410} & \textbf{0.448} & 0.432 & 0.471 & \textbf{0.397} & \textbf{0.436} & 0.438 & 0.465 & \textbf{0.439} & \textbf{0.452} \\
ETTh2-AVG & 0.344 & 0.391 & \textbf{0.321} & \textbf{0.376} & 0.392 & 0.422 & \textbf{0.373} & \textbf{0.409} & 0.379 & 0.426 & \textbf{0.349} & \textbf{0.395} & 0.433 & 0.455 & \textbf{0.396} & \textbf{0.414} \\
\midrule

ETTm1-96  & 0.298 & 0.352 & {0.299} & {0.352} & 0.315 & 0.369 & \textbf{0.310} & \textbf{0.363} & 0.332 & 0.384 & \textbf{0.306} & \textbf{0.358} & 0.359 & 0.391 & \textbf{0.313} & \textbf{0.362} \\
ETTm1-192 & 0.335 & 0.373 & \textbf{0.327} & \textbf{0.370} & 0.349 & 0.388 & \textbf{0.349} & \textbf{0.389} & 0.355 & 0.398 & \textbf{0.340} & \textbf{0.374} & 0.368 & 0.398 & \textbf{0.346} & \textbf{0.383} \\
ETTm1-336 & 0.366 & 0.394 & \textbf{0.354} & \textbf{0.390} & 0.381 & 0.409 & \textbf{0.378} & \textbf{0.406} & 0.386 & 0.416 & \textbf{0.366} & \textbf{0.389} & 0.429 & 0.438 & \textbf{0.378} & \textbf{0.400} \\
ETTm1-720 & 0.420 & 0.421 & \textbf{0.401} & \textbf{0.418} & 0.437 & 0.439 & \textbf{0.431} & \textbf{0.434} & 0.452 & 0.457 & \textbf{0.426} & \textbf{0.425} & 0.477 & 0.474 & \textbf{0.449} & \textbf{0.440} \\
ETTm1-AVG & 0.355 & 0.385 & \textbf{0.345} & \textbf{0.383} & 0.371 & 0.401 & \textbf{0.367} & \textbf{0.398} & 0.381 & 0.414 & \textbf{0.360} & \textbf{0.387} & 0.408 & 0.425 & \textbf{0.372} & \textbf{0.396} \\
\midrule

ETTm2-96  & 0.165 & 0.260 & \textbf{0.163} & \textbf{0.253} & 0.179 & 0.274 & \textbf{0.174} & \textbf{0.265} & 0.192 & 0.285 & \textbf{0.173} & \textbf{0.256} & 0.200 & 0.288 & \textbf{0.181} & \textbf{0.274} \\
ETTm2-192 & 0.219 & 0.298 & \textbf{0.221} & \textbf{0.297} & 0.239 & 0.314 & \textbf{0.240} & \textbf{0.308} & 0.253 & 0.329 & \textbf{0.237} & \textbf{0.297} & 0.274 & 0.337 & \textbf{0.250} & \textbf{0.316} \\
ETTm2-336 & 0.268 & 0.333 & \textbf{0.269} & \textbf{0.327} & 0.309 & 0.356 & \textbf{0.293} & \textbf{0.347} & 0.307 & 0.362 & \textbf{0.294} & \textbf{0.335} & 0.340 & 0.382 & \textbf{0.289} & \textbf{0.340} \\
ETTm2-720 & 0.352 & 0.386 & \textbf{0.347} & \textbf{0.378} & 0.387 & 0.407 & \textbf{0.364} & \textbf{0.391} & 0.380 & 0.412 & \textbf{0.390} & \textbf{0.393} & 0.384 & 0.407 & \textbf{0.380} & \textbf{0.399} \\
ETTm2-AVG & 0.251 & 0.319 & \textbf{0.250} & \textbf{0.314} & 0.279 & 0.338 & \textbf{0.268} & \textbf{0.328} & 0.283 & 0.347 & \textbf{0.274} & \textbf{0.320} & 0.300 & 0.354 & \textbf{0.275} & \textbf{0.332} \\
\midrule

Weather-96  & 0.149 & 0.199 & \textbf{0.146} & \textbf{0.197} & 0.159 & 0.212 & {0.166} & {0.221} & 0.163 & 0.223 & \textbf{0.153} & \textbf{0.208} & 0.176 & 0.234 & \textbf{0.176} & \textbf{0.233} \\
Weather-192 & 0.193 & 0.243 & \textbf{0.191} & \textbf{0.242} & 0.203 & 0.252 & {0.227} & {0.256} & 0.201 & 0.254 & \textbf{0.199} & \textbf{0.249} & 0.219 & 0.270 & {0.220} & {0.271} \\
Weather-336 & 0.240 & 0.281 & \textbf{0.241} & \textbf{0.280} & 0.253 & 0.291 & {0.254} & {0.292} & 0.258 & 0.300 & \textbf{0.253} & \textbf{0.293} & 0.277 & 0.311 & \textbf{0.260} & \textbf{0.295} \\
Weather-720 & 0.312 & 0.334 & \textbf{0.305} & \textbf{0.328} & 0.317 & 0.337 & {0.322} & {0.344} & 0.329 & 0.348 & \textbf{0.330} & \textbf{0.348} & 0.344 & 0.356 & \textbf{0.319} & \textbf{0.341} \\
Weather-AVG & 0.224 & 0.264 & \textbf{0.221} & \textbf{0.262} & 0.233 & 0.273 & {0.242} & {0.278} & 0.238 & 0.281 & \textbf{0.234} & \textbf{0.275} & 0.254 & 0.293 & \textbf{0.244} & \textbf{0.285} \\
\midrule

Electricity-96  & 0.141 & 0.240 & \textbf{0.130} & \textbf{0.225} & 0.135 & 0.233 & \textbf{0.131} & \textbf{0.228} & 0.139 & 0.239 & \textbf{0.131} & \textbf{0.226} & 0.202 & 0.308 & \textbf{0.180} & \textbf{0.283} \\
Electricity-192 & 0.156 & 0.256 & \textbf{0.147} & \textbf{0.241} & 0.155 & 0.253 & \textbf{0.153} & \textbf{0.250} & 0.155 & 0.250 & \textbf{0.149} & \textbf{0.243} & 0.218 & 0.322 & \textbf{0.187} & \textbf{0.292} \\
Electricity-336 & 0.172 & 0.267 & \textbf{0.161} & \textbf{0.256} & 0.169 & 0.267 & \textbf{0.168} & \textbf{0.265} & 0.171 & 0.265 & \textbf{0.164} & \textbf{0.258} & 0.232 & 0.332 & \textbf{0.211} & \textbf{0.311} \\
Electricity-720 & 0.208 & 0.299 & \textbf{0.195} & \textbf{0.288} & 0.204 & 0.301 & \textbf{0.189} & \textbf{0.284} & 0.208 & 0.300 & \textbf{0.200} & \textbf{0.293} & 0.299 & 0.375 & \textbf{0.244} & \textbf{0.335} \\
Electricity-AVG & 0.169 & 0.266 & \textbf{0.158} & \textbf{0.253} & 0.166 & 0.264 & \textbf{0.160} & \textbf{0.257} & 0.168 & 0.264 & \textbf{0.161} & \textbf{0.255} & 0.238 & 0.334 & \textbf{0.206} & \textbf{0.305} \\
\bottomrule
\end{tabular}
}
\end{table*}
\begin{table*}[t]
\centering
\scriptsize
\setlength{\tabcolsep}{3pt}
\renewcommand{\arraystretch}{1.08}
\caption{Long-term multivariate forecasting results on six datasets. The table reports MSE and MAE for forecasting horizons $H \in \{96,192,336,720\}$. The input sequence length is fixed to 720 for all methods. In each row, the best and second-best results for each metric are highlighted in \textcolor{red}{red} and \textcolor{blue}{blue}, respectively. In case of ties, priority is given to the earlier model in the table. The results of the other baseline models are directly obtained from PhaseFormer~\cite{khan2025phaseformer} and TimeBase~\cite{huangtimebase}.}
\label{tab:full_results_updated}
\resizebox{\textwidth}{!}{
\begin{tabular}{llcccccccccccccc}
\toprule
\multirow{2}{*}{Dataset} & \multirow{2}{*}{$H$}
& \multicolumn{2}{c}{DeltaTSF}
& \multicolumn{2}{c}{PhaseFormer}
& \multicolumn{2}{c}{PatchTST}
& \multicolumn{2}{c}{iTransformer}
& \multicolumn{2}{c}{TimeBase}
& \multicolumn{2}{c}{PatchMLP}
& \multicolumn{2}{c}{DLinear} \\
\cmidrule(lr){3-4}\cmidrule(lr){5-6}\cmidrule(lr){7-8}\cmidrule(lr){9-10}\cmidrule(lr){11-12}\cmidrule(lr){13-14}\cmidrule(lr){15-16}
& & MSE & MAE & MSE & MAE & MSE & MAE & MSE & MAE & MSE & MAE & MSE & MAE & MSE & MAE \\
\midrule

\multirow{5}{*}{ETTh1}
& 96  & \textcolor{blue}{\textbf{0.362}} & 0.395 & \textcolor{red}{\textbf{0.359}} & \textcolor{red}{\textbf{0.382}} & 0.377 & 0.408 & 0.389 & 0.421 & 0.365 & \textcolor{blue}{\textbf{0.387}} & 0.417 & 0.440 & 0.378 & 0.402 \\
& 192 & \textcolor{red}{\textbf{0.396}} & 0.416 & \textcolor{blue}{\textbf{0.397}} & \textcolor{red}{\textbf{0.404}} & 0.413 & 0.431 & 0.424 & 0.446 & 0.403 & \textcolor{blue}{\textbf{0.409}} & 0.454 & 0.465 & 0.415 & 0.425 \\
& 336 & \textcolor{red}{\textbf{0.405}} & \textcolor{blue}{\textbf{0.424}} & 0.425 & 0.424 & 0.436 & 0.446 & 0.456 & 0.469 & \textcolor{blue}{\textbf{0.409}} & \textcolor{red}{\textbf{0.419}} & 0.514 & 0.503 & 0.449 & 0.449 \\
& 720 & \textcolor{red}{\textbf{0.420}} & \textcolor{red}{\textbf{0.448}} & \textcolor{blue}{\textbf{0.431}} & 0.450 & 0.455 & 0.475 & 0.545 & 0.532 & 0.440 & \textcolor{blue}{\textbf{0.448}} & 0.548 & 0.531 & 0.507 & 0.517 \\
& AVG & \textcolor{red}{\textbf{0.396}} & 0.421 & \textcolor{blue}{\textbf{0.403}} & \textcolor{red}{\textbf{0.415}} & 0.420 & 0.440 & 0.454 & 0.467 & 0.404 & \textcolor{blue}{\textbf{0.416}} & 0.483 & 0.485 & 0.437 & 0.448 \\
\midrule

\multirow{5}{*}{ETTh2}
& 96  & \textcolor{red}{\textbf{0.269}} & \textcolor{red}{\textbf{0.333}} & \textcolor{blue}{\textbf{0.275}} & \textcolor{blue}{\textbf{0.338}} & 0.276 & 0.339 & 0.305 & 0.361 & 0.292 & 0.350 & 0.324 & 0.382 & 0.294 & 0.360 \\
& 192 & \textcolor{red}{\textbf{0.330}} & \textcolor{red}{\textbf{0.375}} & 0.341 & \textcolor{blue}{\textbf{0.376}} & 0.342 & 0.385 & 0.405 & 0.421 & \textcolor{blue}{\textbf{0.339}} & 0.387 & 0.381 & 0.418 & 0.412 & 0.437 \\
& 336 & \textcolor{red}{\textbf{0.316}} & \textcolor{red}{\textbf{0.377}} & 0.369 & \textcolor{blue}{\textbf{0.405}} & \textcolor{blue}{\textbf{0.364}} & 0.405 & 0.411 & 0.436 & 0.394 & 0.420 & 0.408 & 0.440 & 0.471 & 0.478 \\
& 720 & \textcolor{red}{\textbf{0.368}} & \textcolor{red}{\textbf{0.417}} & 0.402 & 0.436 & \textcolor{blue}{\textbf{0.395}} & \textcolor{blue}{\textbf{0.434}} & 0.448 & 0.470 & 0.400 & 0.448 & 0.454 & 0.469 & 0.748 & 0.609 \\
& AVG & \textcolor{red}{\textbf{0.321}} & \textcolor{red}{\textbf{0.376}} & 0.347 & \textcolor{blue}{\textbf{0.389}} & \textcolor{blue}{\textbf{0.344}} & 0.391 & 0.392 & 0.422 & 0.356 & 0.401 & 0.392 & 0.427 & 0.481 & 0.471 \\
\midrule

\multirow{5}{*}{ETTm1}
& 96  & 0.299 & 0.352 & \textcolor{red}{\textbf{0.293}} & \textcolor{red}{\textbf{0.344}} & \textcolor{blue}{\textbf{0.298}} & 0.352 & 0.315 & 0.369 & 0.311 & 0.351 & 0.319 & 0.367 & 0.314 & \textcolor{blue}{\textbf{0.350}} \\
& 192 & \textcolor{blue}{\textbf{0.327}} & \textcolor{blue}{\textbf{0.370}} & \textcolor{red}{\textbf{0.323}} & \textcolor{red}{\textbf{0.361}} & 0.335 & 0.373 & 0.349 & 0.388 & 0.338 & 0.371 & 0.353 & 0.385 & 0.347 & 0.381 \\
& 336 & \textcolor{red}{\textbf{0.354}} & 0.390 & \textcolor{blue}{\textbf{0.358}} & \textcolor{red}{\textbf{0.381}} & 0.366 & 0.394 & 0.381 & 0.409 & 0.364 & \textcolor{blue}{\textbf{0.386}} & 0.381 & 0.401 & 0.367 & 0.387 \\
& 720 & \textcolor{red}{\textbf{0.401}} & 0.418 & \textcolor{blue}{\textbf{0.412}} & \textcolor{red}{\textbf{0.410}} & 0.420 & 0.421 & 0.437 & 0.439 & 0.413 & \textcolor{blue}{\textbf{0.414}} & 0.436 & 0.432 & 0.415 & 0.415 \\
& AVG & \textcolor{red}{\textbf{0.345}} & 0.383 & \textcolor{blue}{\textbf{0.347}} & \textcolor{red}{\textbf{0.374}} & 0.355 & 0.385 & 0.371 & 0.401 & 0.357 & \textcolor{blue}{\textbf{0.381}} & 0.372 & 0.396 & 0.361 & 0.383 \\
\midrule

\multirow{5}{*}{ETTm2}
& 96  & \textcolor{blue}{\textbf{0.163}} & \textcolor{red}{\textbf{0.253}} & 0.163 & \textcolor{blue}{\textbf{0.256}} & 0.165 & 0.260 & 0.179 & 0.274 & \textcolor{red}{\textbf{0.162}} & 0.256 & 0.186 & 0.272 & 0.163 & 0.257 \\
& 192 & 0.221 & 0.297 & \textcolor{blue}{\textbf{0.219}} & \textcolor{red}{\textbf{0.293}} & 0.219 & 0.298 & 0.239 & 0.314 & \textcolor{red}{\textbf{0.218}} & \textcolor{blue}{\textbf{0.293}} & 0.250 & 0.315 & 0.223 & 0.304 \\
& 336 & \textcolor{blue}{\textbf{0.269}} & \textcolor{blue}{\textbf{0.327}} & 0.269 & \textcolor{red}{\textbf{0.326}} & \textcolor{red}{\textbf{0.268}} & 0.333 & 0.309 & 0.356 & 0.270 & 0.328 & 0.298 & 0.347 & 0.291 & 0.355 \\
& 720 & \textcolor{red}{\textbf{0.347}} & \textcolor{red}{\textbf{0.378}} & \textcolor{blue}{\textbf{0.351}} & \textcolor{blue}{\textbf{0.379}} & 0.352 & 0.386 & 0.387 & 0.407 & 0.352 & 0.380 & 0.380 & 0.398 & 0.407 & 0.433 \\
& AVG & \textcolor{red}{\textbf{0.250}} & \textcolor{red}{\textbf{0.314}} & \textcolor{blue}{\textbf{0.251}} & \textcolor{blue}{\textbf{0.314}} & 0.251 & 0.319 & 0.279 & 0.338 & 0.251 & 0.314 & 0.279 & 0.333 & 0.271 & 0.337 \\
\midrule

\multirow{5}{*}{Weather}
& 96  & \textcolor{red}{\textbf{0.146}} & \textcolor{blue}{\textbf{0.197}} & 0.148 & \textcolor{red}{\textbf{0.195}} & 0.149 & 0.199 & 0.159 & 0.212 & \textcolor{blue}{\textbf{0.146}} & 0.198 & 0.152 & 0.205 & 0.174 & 0.242 \\
& 192 & \textcolor{blue}{\textbf{0.191}} & 0.242 & 0.193 & \textcolor{red}{\textbf{0.237}} & 0.193 & 0.243 & 0.203 & 0.252 & \textcolor{red}{\textbf{0.185}} & \textcolor{blue}{\textbf{0.241}} & 0.197 & 0.247 & 0.215 & 0.277 \\
& 336 & \textcolor{blue}{\textbf{0.241}} & \textcolor{blue}{\textbf{0.280}} & 0.242 & \textcolor{red}{\textbf{0.278}} & \textcolor{red}{\textbf{0.240}} & 0.281 & 0.253 & 0.291 & 0.263 & 0.281 & 0.248 & 0.287 & 0.262 & 0.319 \\
& 720 & \textcolor{red}{\textbf{0.305}} & \textcolor{red}{\textbf{0.328}} & \textcolor{blue}{\textbf{0.309}} & 0.332 & 0.312 & 0.334 & 0.317 & 0.337 & 0.314 & \textcolor{blue}{\textbf{0.331}} & 0.324 & 0.340 & 0.319 & 0.359 \\
& AVG & \textcolor{red}{\textbf{0.221}} & \textcolor{blue}{\textbf{0.262}} & \textcolor{blue}{\textbf{0.223}} & \textcolor{red}{\textbf{0.261}} & 0.224 & 0.264 & 0.233 & 0.273 & 0.227 & 0.263 & 0.230 & 0.270 & 0.243 & 0.299 \\
\midrule

\multirow{5}{*}{Electricity}
& 96  & \textcolor{blue}{\textbf{0.130}} & \textcolor{blue}{\textbf{0.225}} & \textcolor{red}{\textbf{0.129}} & \textcolor{red}{\textbf{0.221}} & 0.141 & 0.240 & 0.135 & 0.233 & 0.139 & 0.231 & 0.141 & 0.243 & 0.141 & 0.244 \\
& 192 & \textcolor{red}{\textbf{0.147}} & \textcolor{blue}{\textbf{0.241}} & \textcolor{blue}{\textbf{0.148}} & \textcolor{red}{\textbf{0.238}} & 0.156 & 0.256 & 0.155 & 0.253 & 0.153 & 0.245 & 0.163 & 0.265 & 0.155 & 0.258 \\
& 336 & \textcolor{red}{\textbf{0.161}} & \textcolor{red}{\textbf{0.256}} & \textcolor{blue}{\textbf{0.165}} & \textcolor{blue}{\textbf{0.257}} & 0.172 & 0.267 & 0.169 & 0.267 & 0.169 & 0.262 & 0.186 & 0.290 & 0.170 & 0.275 \\
& 720 & \textcolor{red}{\textbf{0.195}} & \textcolor{blue}{\textbf{0.288}} & \textcolor{blue}{\textbf{0.201}} & \textcolor{red}{\textbf{0.285}} & 0.208 & 0.299 & 0.204 & 0.301 & 0.207 & 0.294 & 0.221 & 0.320 & 0.209 & 0.309 \\
& AVG & \textcolor{red}{\textbf{0.158}} & \textcolor{blue}{\textbf{0.253}} & \textcolor{blue}{\textbf{0.161}} & \textcolor{red}{\textbf{0.250}} & 0.169 & 0.266 & 0.166 & 0.264 & 0.167 & 0.258 & 0.178 & 0.280 & 0.169 & 0.272 \\
\bottomrule
\end{tabular}
}
\end{table*}

\end{document}